\documentclass{preprint}

\usepackage{amsmath,amssymb,amsfonts}
\usepackage{algorithmic}
\usepackage{algorithm}
\usepackage{graphicx}
\usepackage{xcolor}
\usepackage{booktabs}
\usepackage{tabularx}
\usepackage{mdframed}
\usepackage{subcaption}
\usepackage{listings}
\usepackage{colortbl}
\usepackage[numbers,comma,sort&compress]{natbib} 
\usepackage[colorlinks=true]{hyperref}

\definecolor{Stone0}{HTML}{F6F3EB}

\newcounter{prompt}
\newenvironment{prompt}[1][]{
\refstepcounter{prompt}
\begin{mdframed}[
innertopmargin=2pt, 
innerbottommargin=2pt,
innerleftmargin=2pt, 
innerrightmargin=2pt,
frametitle={\textmd{Prompt \theprompt: #1}},
frametitlerule=true,
frametitleaboveskip=2pt,
frametitlebelowskip=2pt,
backgroundcolor=Stone0]%
}
{%
\par%
\end{mdframed}%
}

\colorlet{punct}{red!60!black}
\definecolor{background}{HTML}{EEEEEE}
\definecolor{delim}{RGB}{20,105,176}
\colorlet{numb}{magenta!60!black}
\lstdefinelanguage{json}{
    basicstyle=\ttfamily\footnotesize,
    numberstyle=\scriptsize,
    stepnumber=1,
    numbersep=8pt,
    showstringspaces=false,
    breaklines=true,
    literate=
     *{0}{{{\color{numb}0}}}{1}
      {1}{{{\color{numb}1}}}{1}
      {2}{{{\color{numb}2}}}{1}
      {3}{{{\color{numb}3}}}{1}
      {4}{{{\color{numb}4}}}{1}
      {5}{{{\color{numb}5}}}{1}
      {6}{{{\color{numb}6}}}{1}
      {7}{{{\color{numb}7}}}{1}
      {8}{{{\color{numb}8}}}{1}
      {9}{{{\color{numb}9}}}{1}
      {:}{{{\color{punct}{:}}}}{1}
      {,}{{{\color{punct}{,}}}}{1}
      {\{}{{{\color{delim}{\{}}}}{1}
      {\}}{{{\color{delim}{\}}}}}{1}
      {[}{{{\color{delim}{[}}}}{1}
      {]}{{{\color{delim}{]}}}}{1},
}

\lstdefinelanguage{prompt}{
    breaklines=true,
    breakindent=0pt,
    basicstyle=\ttfamily\footnotesize,
    identifierstyle=,
    columns=fullflexible
}

\title{Using reasoning LLMs to extract SDOH events from clinical notes}

\author[1]{Ertan Dogan}
\author[1]{Kunyu Yu}
\author[1,*]{Yifan Peng}
\affil[1]{Department of Population Health Sciences, Weill Cornell Medicine, New York, USA}
\affil[*]{Corresponding author(s). Email(s): \url{yip4002@med.cornell.edu}}

\begin{document}

\maketitle

\begin{abstract}
Social Determinants of Health (SDOH) refer to environmental, behavioral, and social conditions that influence how individuals live, work, and age. SDOH have a significant impact on personal health outcomes,  and their systematic identification and management can yield substantial improvements in patient care. However, SDOH information is predominantly captured in unstructured clinical notes within electronic health records, which limits its direct use as machine-readable entities. To address this issue, researchers have employed Natural Language Processing (NLP) techniques using pre-trained BERT-based models, demonstrating promising performance but requiring sophisticated implementation and extensive computational resources. In this study, we investigated prompt engineering strategies for extracting structured SDOH events utilizing LLMs with advanced reasoning capabilities. Our method consisted of four modules: 1) developing concise and descriptive prompts integrated with established guidelines, 2) applying few-shot learning with carefully curated examples, 3) using a self-consistency mechanism to ensure robust outputs, and 4) post-processing for quality control. Our approach achieved a micro-F1 score of 0.866, demonstrating competitive performance compared to the leading models. The results demonstrated that LLMs with reasoning capabilities are effective solutions for SDOH event extraction, offering both implementation simplicity and strong performance.
\end{abstract}


\section{Introduction}
Social Determinants of Health (SDOH) are the conditions in which people live, work, and organize their lives, ultimately influencing their health outcomes~\cite{healthyPeople20302025social}. These determinants encompass both protective factors that reduce health risks, such as access to healthcare and safe environments, and risk factors that can lead to adverse health outcomes, including poverty and substance abuse. SDOHs are traditionally grouped into five domains: economic stability, access to and quality of education, access to and quality of healthcare, neighborhood and built environment, and social and community context \cite{healthyPeople20302025social}. Growing evidence indicates that SDOH have a significant impact on population health and quality of life, with significant effects on both physical and mental health \cite{healthyPeople20302025social, li2024realizing}. 

Due to their importance for individual well-being and public health, researchers seek to ensure that these factors are systematically incorporated into patient diagnosis and clinical decision-making processes. Since most SDOH data exists within unstructured clinical notes generated through natural patient-provider interactions rather than in structured, machine-readable formats \cite{lybarger20232022, yu2022assessing}, specialized processing methods are required to extract and standardize this valuable information into usable formats. Consequently, researchers have developed strategies using Natural Language Processing (NLP) methods, including rule-based approaches and pre-trained language models (LMs) \cite{yu2024identifying, gabriel2024development, lybarger2023leveraging, rawat2022investigation}. 

\begin{figure}[t]
    \centering
    \begin{subfigure}{.45\linewidth}
    \centering
    \includegraphics[width=\linewidth, clip, trim=0in .2in 3.6in .2in]{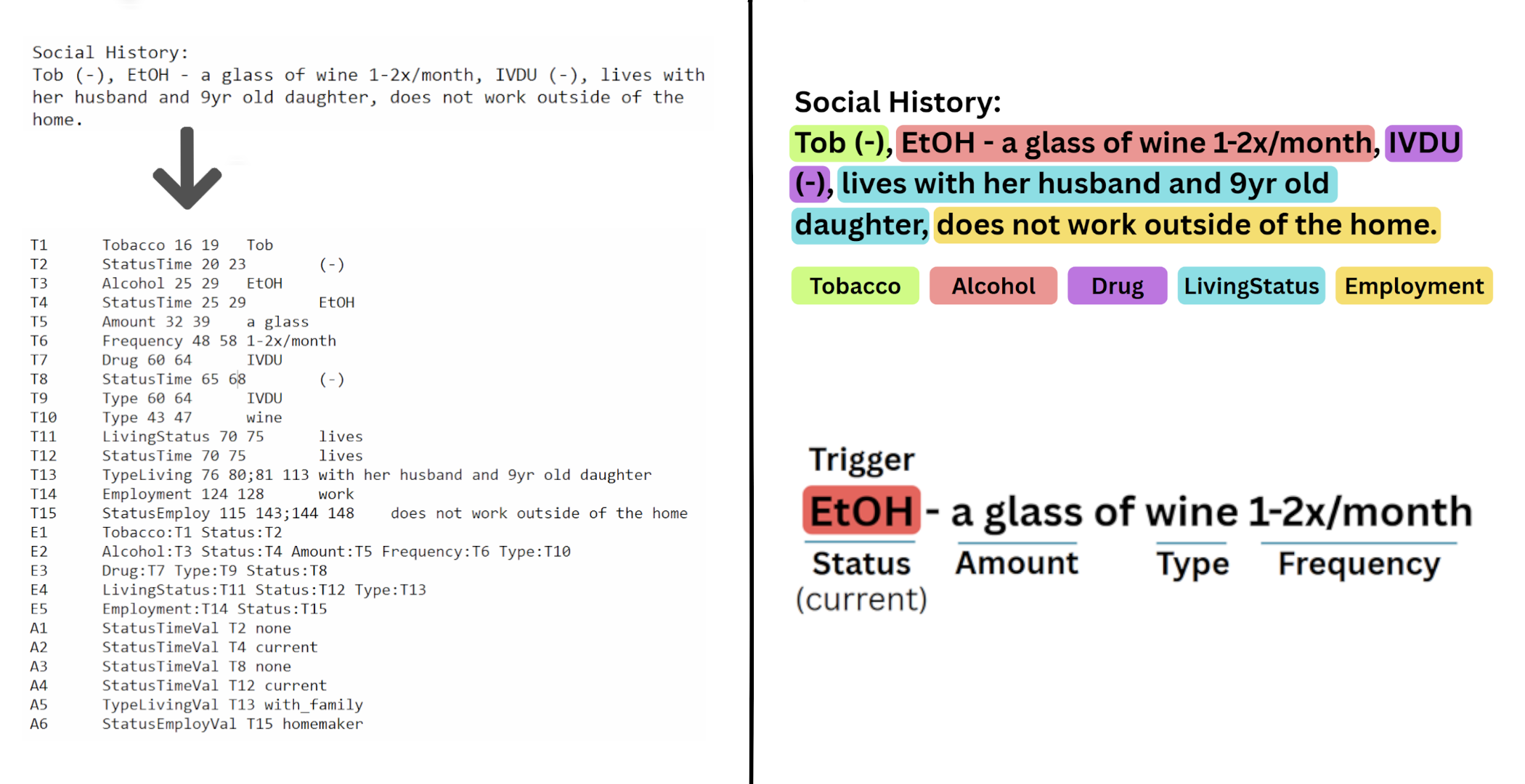}
    \caption{\label{fig:brat}}
    \end{subfigure}
    \begin{subfigure}{.45\linewidth}
    \centering
    \includegraphics[width=.7\linewidth, clip, trim=3.5in .7in 0 .3in]{Annotations.png}
    \caption{\label{fig:vis}}
    \end{subfigure}
    \caption{A sample of SDOH annotations in the 2022 n2c2/UW shared task.}
    \label{fig:annotations}
\end{figure}

A recent effort is the 2022 n2c2/UW shared task on extracting SDOH. This competition scored participating research teams on the extraction of five categories of SDOH events: tobacco use, alcohol use, drug use, living status, and employment status \cite{lybarger20232022}. Teams were tasked with extracting events as combinations of labeled text spans and set values that provided comprehensive and accurate descriptions of patient SDOH events. An example of annotations is shown in Figure~\ref{fig:annotations}. 

The top performing teams in the 2022 n2c2/UW shared task leveraged LMs pre-trained for Named Entity Recognition (NER), with most approaches involving fine-tuning BERT-based models for SDOH extraction \cite{devlin2019bert, lybarger2023leveraging}.
While these methods achieved strong performance, their reliance on extensive pretraining and large annotated datasets created barriers to broader adoption. In addition, such models require further fine-tuning to recognize SDOH beyond the shared task scope, a resource-intensive process often hindered by limited data on rare SDOH events.
Building on this challenge, subsequent studies used the annotated dataset with advanced large language models (LLMs), including OpenAI's ChatGPT-4 and Meta's LLaMa \cite{yu2024identifying,gabriel2024development}. While these efforts achieved notable results, they did not match the performance of fine-tuned BERT-based models on event extraction. The LLMs used were general-domain models without domain-specific fine-tuning for medical text analysis, resulting in inferior performance compared to task-specific BERT models that benefit from substantial training data \cite{gabriel2024development}. For less common SDOH, LLMs tended to perform closer to BERT-based models. The recent emergence of LLMs with enhanced reasoning capabilities, such as OpenAI's o4-mini and Deepseek's R1, offers new opportunities to advance SDOH extraction.

In this study, we propose optimizing LLM prompting strategies to provide precise instructions and guidelines while leveraging the advanced reasoning capabilities of next-generation LLMs. Specifically, our objective is two-fold: (1) to investigate various prompting strategies for SDOH extraction using reasoning LLMs, and (2) to identify an intuitive querying approach that enables reasoning-capable LLMs to effectively identify and extract SDOH information from patient records in a structured format readily applicable to clinical decision-making workflows.

We evaluated our approach on the 2022 n2c2/UW SDOH shared task using three reasoning LLMs, including OpenAI's o4-mini, Gemini 2.5 Flash, and Llama-3.1-8B. Experiments demonstrated that our strategy using OpenAI's o4-mini model yielded results competitive with those of the top-performing teams in the n2c2/UW shared task. Additionally, our strategy achieved higher precision than all leading teams. 

\section{Related Work}
Recent studies have investigated the application of LLMs to health-related topics \cite{consoli2025sdoh, guevara2024large, conway2019moonstone, bejan2018mining, han2022classifying, feller2020detecting, patra2021extracting, romanowski2023extracting, richie2023extracting, pethani2025extracting, chen2025extraction, wang2025extracting}. For example, Guevara et al.\cite{guevara2024large} and Consoli et al.\cite{consoli2025sdoh} applied LLMs to clinical texts to identify SDOH information and evaluated ChatGPT-family models (GPT-3.5, GPT-4) using few-shot prompt strategies. These studies demonstrated results comparable to those of traditional language-model-based strategies, but primarily focused on classifying the presence of SDOH events rather than performing span-level extraction of SDOH events. In contrast, our work investigates the use of LLMs for span-level SDOH extraction, capturing detailed attributes such as event status, history, and the amount (for substance use).

In parallel, LLMs with reasoning capabilities have been enhanced in models such as OpenAI and DeepSeek, incorporating techniques such as chain-of-thought prompting to improve performance on tasks requiring logic inference and analytical reasoning \cite{shao2024deepseekmath, lambert2024tulu, tordjman2025comparative, wang2023selfconsistencyimproveschainthought}. For example, Tordjman et al. \cite{tordjman2025comparative} applied models such as OpenAI's O1 and Deepseek's R1 to text-based diagnosis and summarization and achieved promising results. However, these studies did not explore SDOH analysis or extraction. Building on this line of work, we apply reasoning-capable LLMs, including OpenAI’s o4-mini, Gemini 2.5 Flash, and Llama-3.1-8B, to SDOH extraction at the span level from clinical texts. We also apply self-consistency techniques \cite{wang2023selfconsistencyimproveschainthought} alongside chain-of-thought reasoning to improve extraction accuracy and robustness for SDOH events.

\section{Materials and Methods}


\subsection{Dataset}

The Social History Annotated Corpus (SHAC) corpus \cite{lybarger2021annotating}, used in the 2022 n2c2/UW SDOH challenge, is a publicly available, deidentified health records corpus containing 4,480 annotated social history sections (Table~\ref{tab:dataset}). SHAC comprises two data sources: MIMIC-III and the University of Washington (UW), each divided into training, testing, and development subsets. 
The annotations were in the Brat standoff format \cite{stenetorp2012brat}, a human- and machine-readable annotation format that involves labeled text spans which can be connected into events. The data was annotated for five SDOH categories: living status, employment status, tobacco use, drug use, and alcohol use. Each event includes the relevant trigger and all present span-based and labeled arguments. 

For each extracted event, the annotated arguments include trigger, status, type, duration, history, frequency (applicable only to alcohol use, drug use, and tobacco use events), and amount (applicable only to alcohol use, drug use, and tobacco use events). Further details are provided in Table 1 of Lybarger et al. \cite{lybarger20232022}.


\begin{table}[t]
\centering
\caption{Characteristics of the dataset.}
\label{tab:dataset}
\small
\begin{tabular}{lrrr}
\toprule
     &  MIMIC-III & UW & Total\\
\midrule
Clinical notes  & 1,880 & 2,600 & 4,480\\
Avg number of words & 39.15 & 36.32 & 37.52\\
Annotations\\
~~Alcohol & 1,424 & 1,870 & 3,294\\
~~Drug & 940 & 1,913 & 2,853\\
~~Employment & 1,012 & 826 & 1,838\\
~~Living Status & 1,113 & 1,567 & 2,680\\
~~Tobacco & 1,401 & 1,872 & 3,273\\
\bottomrule
\end{tabular}
\end{table}

\subsection{Task Definition}

The task was to extract SDOH from clinical notes describing patients' social histories. We followed the annotation framework of the SHAC corpus, which represents each SDOH as a structured event. Each event is denoted by a trigger and several span-based and labeled arguments. These arguments are: status, type, duration, history, frequency, amount, and method. The rules governing event extraction are described in the SHAC Annotation Guidelines \cite{lybarger2021annotating}. 

Figure \ref{fig:brat} shows an excerpt from the clinical record, followed by its machine-readable format listing all identified SDOH events and their arguments. Figure \ref{fig:vis} presents the same information in a more visually appealing format. The different colors represent distinct event types: alcohol use, drug use, employment status, living status, and tobacco use. We also show the individual annotations for the alcohol event trigger and its arguments: status (current), amount (a glass), type (wine), and frequency (1-2x/month). These excerpts highlight the information density of such notes. A successful evaluation of this task requires the precise identification of all these elements.

\subsection{Model Developement}

\begin{figure}[t]
    \centering
    \includegraphics[width=.7\linewidth]{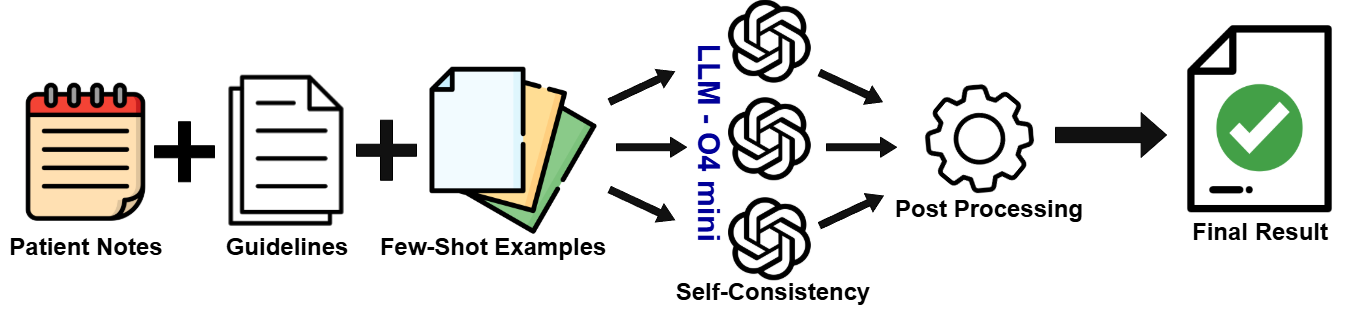}
    \caption{Our proposed SDOH extraction pipeline.}
    \label{fig:pipeline}
    \vspace{-1em}
\end{figure}


%


Our approach contained four key modules (Figure \ref{fig:pipeline}): 1) prompts with the SHAC annotation guidelines, 2) optimizing few-shot prompting, 3) self-consistency, and 4) post-processing.  

\subsubsection{The Prompt with Guidelines}

For our study, we utilized three models with reasoning capabilities, namely OpenAI's o4-mini, Gemini 2.5 Flash, and Llama-3.1-8B. Our prompt relied on the SHAC annotation guidelines to instruct the model. The annotation guidelines are uploaded to the model, providing detailed instructions for extracting each SDOH type and its corresponding argument. The model generates all SDOH annotations at once. The annotations are in JavaScript Object Notation (JSON) and are then converted to the Brat standoff format. The prompt used to provide the annotation guidelines directly to the model is shown in Prompt \ref{prompt:guidelines}.

\begin{prompt}[Annotating SDOH]
\label{prompt:guidelines}
\begin{lstlisting}[language=prompt]
Using the uploaded annotation guidelines for clinical notes, annotate the following clinical note from a doctor about a patient for the social determinants of health present in the notes. Output a list of events formatted in JSON format like this:
\end{lstlisting}

\begin{lstlisting}[language=json]
[{
  "sdoh": "[Employment|LivingStatus|Alcohol|Tobacco|Drug]",
  "trigger": (start_index, end_index, "text"),
  "status": (start_index, end_index, "text", "value") or null,
  "duration": (start_index, end_index, "text") or null,
  "history": (start_index, end_index, "text") or null,
  "type": (start_index, end_index, "text", "value") or null,
  "amount": (start_index, end_index, "text") or null,
  "frequency": (start_index, end_index, "text") or null,
  "method": (start_index, end_index, "text") or null
}]
\end{lstlisting}
\begin{lstlisting}[language=prompt]
Only include a field if it is present in the notes. Check for the trigger (required), status (required), type (required for LivingStatus), duration, frequency, history, method, and amount. 

Sometimes the input text is blank, or no events are found. Simply return an empty list of annotations.

Only return the output list. Add no comments, citations, or extra notes.

Here are the notes to annotate:
\end{lstlisting}
\begin{lstlisting}[language=prompt, escapechar=\%]
    A note sample is shown in Figure %\ref{fig:brat}%
\end{lstlisting}
\end{prompt}

\subsubsection{Few-shot Prompting}

For few-shot prompting, we provided the model with 50 fully annotated examples drawn from the MIMIC-III and UW training datasets, formatted in JSON in the same way as the model is instructed to format its response. The format of the examples we provided to the model is shown in Prompt \ref{prompt:example}.

\begin{prompt}[Examples provided to the model]
\label{prompt:example}
\begin{lstlisting}[language=prompt]
Notes: SOCIAL HISTORY:  Patient lives alone in [**Hospital1 **]. She has a daughter who lives five minutes away.  The patient does all of her own cooking and cleaning.  She has no history of alcohol abuse.  She quit smoking 30 years ago. She is a widow.

Annotations:
\end{lstlisting}
\begin{lstlisting}[language=json]
[
  {
    "sdoh": "LivingStatus",
    "trigger": (25, 30, "lives"),
    "status": (25, 30, "lives", "current"),
    "type": (31, 56, "alone in [**Hospital1 **]", "with_others")},
  {
    "sdoh": "Tobacco",
    "trigger": (210, 217, "smoking"),
    "status": (205, 209, "quit", "past"),
    "history": (218, 230, "30 years ago")},
  {
    "sdoh": "Alcohol",
    "trigger": (185, 198, "alcohol abuse"),
    "status": (171, 181, "no history", "none")}
]
\end{lstlisting}
\end{prompt}

\subsubsection{Self-consistency}
We then implemented self-consistency to improve the reliability of predictions. The model was prompted three times for each clinical note, and a majority vote was used to determine the events that were most consistent across the notes. This module mitigates the variability in model outputs and improves F1 performance. The self-consistency algorithm is shown in Algorithm \ref{alg:consistency}.

\subsubsection{Post-processing}

Post-processing was applied to the pipeline outputs to correct errors in character positions that the model would occasionally label inaccurately without a traceable reason. In addition, events were removed if they (1) lacked required arguments, such as a missing ``status'' value, which is required for all extracted SDOH events, or (2) contained invalid values, such as a `status' of ``unknown", whereas valid options are ``past", ``current", or ``none".

\begin{algorithm}[t]
\caption{Self-Consistency Algorithm}
\label{alg:consistency}
\begin{algorithmic}
\STATE $responses \gets []$
\FORALL{$sdoh \in sdoh\_types$}
    \STATE $responses\_per\_sdoh \gets []$
    \FOR{$i = 1, \dots, 3$}
        \STATE Prompt model for SDOH event extraction for $sdoh$
        \STATE Add model response to $responses\_per\_sdoh$
    \ENDFOR
    \STATE Add $responses\_per\_sdoh$ to $responses$
\ENDFOR
\FORALL{$sdoh \in responses$}
    \STATE Determine the most common response for $sdoh$
\ENDFOR
\STATE Compile most common responses to a single annotation
\end{algorithmic}
\end{algorithm}


\subsection{Experimental Settings}

We calculated results by extracting all SDOH events from the clinical notes, including the event trigger and all associated arguments. 

We evaluated the accuracy of span-based events as True Positives when there was overlap between the text extracted by the model and the text in the dataset's annotations. The overlap is calculated by comparing the character positions of the model's extracted text with those of the annotated text. We evaluated the accuracy of value-based events as True Positives for exact matches between the model's classified value and the value in the dataset's annotations. Positive and Negative Values for an attribute were based on whether the model provided a value for that attribute. We reported precision, recall, and Micro-F1.

\section{Results}
\subsection{Overall Results}

Table \ref{tab:results} shows that our method with o4-mini achieved a micro-F1 of 0.866 on the entire testing set. This ranks among the top performers of the n2c2/UW shared task. Also, Gemini 2.5 Flash achieved a micro-F1 of 0.825, demonstrating comparable performance at substantially lower cost. These results showed that our method is capable of accurately annotating SDOH events without task-specific fine-tuning, relying solely on the models' inherent reasoning capabilities.

We also calculated precision, recall, and micro-F1 for each SDOH category in our annotations. O4-mini performed best on Alcohol events (micro-F1: 0.904, precision: 0.918, recall: 0.891), followed by Tobacco events (micro-F1: 0.892, precision: 0.933, recall: 0.854). Gemini 2.5 Flash showed a slightly different pattern, achieving the highest performance on Living Status events with a micro-F1 of 0.910. Notably, both models achieved the lowest performance on Drug events with micro-F1 scores of 0.805 for o4-mini and 0.715 for Gemini 2.5 Flash. This is likely due to the complex annotation scheme for Drug events, which encompasses multiple trigger phrases, methods, and types that challenge the models' extraction capabilities.

In contrast to o4-mini and Gemini 2.5 Flash, Llama-3.1-8B exhibited relatively lower performance with an overall micro-F1 of 0.591, as shown in Supplementary Table \ref{tab:results_appendix}. We observed two key failure modes: inconsistent adherence to the structured SDOH output format and mistaken extraction of information from a few-shot examples rather than the target clinical note. This performance gap suggested that while proprietary reasoning LLMs like o4-mini and Gemini 2.5 Flash can effectively handle complex SDOH extraction tasks, smaller open-source models may require additional optimization to achieve comparable results.


\begin{table}[tb]
\centering
\caption{Comparison of our results with n2c2 shared task teams.} 
\label{tab:results}
\small
\begin{tabular}{lrrr}
\toprule
 & Precision & Recall & Micro-F1\\
\midrule
Microsoft & 0.891 & 0.887 & 0.889\\
Children's Hospital of Philadelphia & 0.874 & 0.888 & 0.881\\
University of Texas at San Antonio & 0.880 & 0.841 & 0.860\\
Philips Research North America & 0.885 & 0.780 & 0.829\\
\midrule
Our method (o4-mini) & \textbf{0.902} & \textbf{0.833} & \textbf{0.866}\\
~~Alcohol & 0.918 & 0.891 & 0.904\\
~~Drug & 0.879 & 0.744 & 0.805\\
~~Employment & 0.889 & 0.786 & 0.844\\
~~Living Status & 0.890 & 0.863 & 0.876\\
~~Tobacco & 0.933 & 0.854 & 0.892\\
\midrule
Our method (gemini-2.5-flash) & 0.833 & 0.817 & 0.825\\
~~Alcohol & 0.782 & 0.823 & 0.802 \\
~~Drug & 0.744 & 0.688 & 0.715 \\
~~Employment & 0.868 & 0.852 & 0.860 \\
~~Living Status & 0.938 & 0.883 & 0.910 \\
~~Tobacco & 0.841 & 0.841 & 0.841 \\
\bottomrule
\end{tabular}
\end{table}

\subsection{Ablation Study}

We performed ablation tests to determine the contribution of each module in the pipeline and to identify potential areas for refinement. 

\subsubsection{Methods of Structuring Guided Prompts}

Our first experiment compared different methods for structuring guided prompts. 
We tested having the LLM annotate each SDOH separately by adding clarifications to the prompt and adjusting the pipeline accordingly. The model still directly used the SHAC annotation guidelines. The prompt is shown at Prompt~\ref{prompt:separatesdoh}.

\begin{prompt}[Annotating Each SDOH Type Separately]
\label{prompt:separatesdoh}
\begin{lstlisting}[language=prompt]
Using the uploaded annotation guidelines for clinical notes, annotate the following clinical note from a doctor about a patient for the social determinants of health present of type {sdoh} in the notes. Output a list of events formatted in JSON format like this:
\end{lstlisting}
\begin{lstlisting}[language=prompt, escapechar=\%]
    Same as in Prompt %\ref{prompt:guidelines}%
\end{lstlisting}

\begin{lstlisting}[language=prompt]
Only include a field if it is present in the notes. Check for the trigger (required), status (required), type (required for LivingStatus), duration, frequency, history, method, and amount. 

Sometimes the input text is blank, or no events are found. Simply return an empty list of annotations.

Extract all of the {sdoh} events in the text. Do not extract any other SDOH type. Do not miss any of the {sdoh} events.

Only return the output list. Add no comments, citations, or extra notes.

Here are the notes to annotate:
\end{lstlisting}

\begin{lstlisting}[language=prompt, escapechar=\%]
    A note sample is shown in Figure %\ref{fig:brat}%
\end{lstlisting}
\end{prompt}

30 examples were randomly selected from the MIMIC-III dataset for this comparison. 
Table \ref{tab:promptmethods} shows that directly providing the full SHAC annotation guidelines led to the best performance, both in asking for the full annotations with one prompt and asking for the annotations of each SDOH type separately in multiple prompts. The single-prompt approach achieved the highest micro-F1 of 0.806, whereas the multi-prompt approach (asking for each SDOH type separately) achieved the highest recall of 0.810. In contrast, the approach that relies on LLM-summarized annotation guidelines achieved the lowest recall among the methods. The LLM-summarized annotation guidelines (shown in Prompt \ref{prompt:llmprompt}) provided a concise description for the model of the structure of annotations and how the SDOH should be extracted. The LLM-summarized annotation guidelines also included a line at the top of the prompt to assign the model a role, a method known as role-prompting, which provided an additional task description and focus to the model. However, the LLM's summary of the annotation guidelines' description of the attributes extracted for each SDOH type omits many details that human annotators would find significant. The SHAC annotation guidelines provide detailed guidance on how to annotate each attribute for each SDOH type. For example, the status attribute is annotated differently for the Employment SDOH type than for any other SDOH type. Because this elaboration is not included in the LLM-summarized annotation guidelines, the model's results are impacted. Based on these findings, we opted to provide the full SHAC annotation guidelines directly to LLM in our experiments.

\begin{prompt}[Prompt with the LLM-summarized annotation guidlines]
\label{prompt:llmprompt}
\begin{lstlisting}[language=prompt]
You are a clinical NLP assistant trained to extract Social Determinants of Health (SDOH) from clinical notes.

An SDOH event is a mention related to:
- **Employment**
- **Living Status**
- **Substance Use**: Alcohol, Tobacco, or Drug use

Each SDOH event must be extracted as a **separate JSON object** with the following structure:
\end{lstlisting}
\begin{lstlisting}[language=prompt, escapechar=\%]
    Same as in Prompt %\ref{prompt:guidelines}%
\end{lstlisting}

\begin{lstlisting}[language=prompt,breakindent=4ex]
### Definitions:
- **Trigger** (required): Phrase that signals the event, such as "works", "drinks", or "lives".
- **Status** (required): Temporal value - e.g., "current", "past", "non
- **Type**: Subcategory - e.g., "nurse", "lives with family", "beer".
- **Duration**: How long the event lasted - e.g., "for 10 years".
- **History**: How long ago the event occurred - e.g., "quit in 2005".
- **Amount**: Quantity - e.g., "2 drinks", "1 pack".
- **Frequency**: How often - e.g., "per day", "weekly".
- **Method**: Only for drug or tobacco (e.g., "injects", "chews tobacco").

Use **character indexes** for each span based on the full note text. Do not merge multiple events; instead, create one JSON object **per SDOH event**.

Example:
\end{lstlisting}

\begin{lstlisting}[language=prompt, escapechar=\%]
    Same as in Prompt %\ref{prompt:example}%
\end{lstlisting}

\begin{lstlisting}[language=prompt]
If a field is null, do not annotate it.

Here are the notes:
\end{lstlisting}

\begin{lstlisting}[language=prompt, escapechar=\%]
    A note sample is shown in Figure %\ref{fig:brat}%
\end{lstlisting}
\end{prompt}


    

\begin{table}[tb]
    \centering
    \caption{Comparing methods of structuring guided prompts on a subset of 30 samples from the MIMIC-III test set.}
    \label{tab:promptmethods}
    \small
    \begin{tabular}{lccc}
\toprule
Prompt Type & Precision & Recall & Micro-F1\\
\midrule
\rowcolor{gray!10}\multicolumn{4}{l}{Single-prompt}\\
SHAC annotation guidelines & 0.806 & 0.806 & 0.806\\
~~Alcohol & 0.830 & 0.846 & 0.838\\
~~Drug & 0.829 & 0.707 & 0.763\\
~~Employment & 0.708 & 0.791 & 0.747\\
~~Living Status & 0.803 & 0.845 & 0.824\\
~~Tobacco & 0.833 & 0.814 & 0.824\\
GPT-generated guidelines & 0.710 & 0.781 & 0.744\\
~~Alcohol & 0.807 & 0.885 & 0.844\\
~~Drug & 0.721 & 0.756 & 0.738\\
~~Employment & 0.592 & 0.674 & 0.630 \\
~~Living Status & 0.571 & 0.690 & 0.625 \\
~~Tobacco & 0.764 & 0.791 & 0.777 \\
\midrule
\rowcolor{gray!10}\multicolumn{4}{l}{Multi-prompt}\\
SHAC annotation guidelines & 0.764 & 0.810 & 0.786\\
~~Alcohol & 0.737 & 0.808 & 0.771\\
~~Drug & 0.706 & 0.585 & 0.640\\
~~Employment & 0.720 & 0.837 & 0.774\\
~~Living Status & 0.852 & 0.897 & 0.874\\
~~Tobacco & 0.791 & 0.837 & 0.814\\
\bottomrule
    \end{tabular}

\end{table}

\subsubsection{Comparison of Few-shot Learning}

We then evaluated the impact of few-shot learning by varying the number of in-context examples in the prompt ($n$) from 0 to 100. This evaluation was conducted using the same 30 examples from the experiment described above. Table \ref{tab:fewshot} shows that 50-shot prompting achieved the highest performance with a micro-F1 of 0.881, and 30-shot prompting achieved a lower but comparable score of 0.872. Additionally, 50-shot prompting significantly outperformed the 0-shot setting (micro-F1 = 0.786), highlighting the importance of including examples in the prompt within our pipeline.

\subsubsection{Contribution of Self-consistency and Post-processing}

Finally, we assessed the impact of self-consistency and post-processing, using a new subset of 30 instances from both the MIMIC-III and UW test sets. Table \ref{tab:selfconsistency} shows that these additions significantly increased performance, together bringing the micro-F1 from 0.842 to 0.929. Using self-consistency alone improved micro-F1 by 0.063, bringing it to 0.903. Post-processing further improved micro-F1 by 0.026, raising it from 0.903 to 0.929. Row 2 shows the model's results after only adding self-consistency to the pipeline, and row 3 displays the results with both self-consistency and post-processing. These experiments also include few-shot prompting in their pipelines. The whole pipeline is as follows. First, few-shot prompting is used to provide an initial prompt to the model. The model is prompted three times to assess self-consistency. From the three sets of annotations provided by the model, a majority vote determines the SDOH events included in the final annotation, as well as the attributes present in each SDOH event. We designed a compilation function to execute this process. Lastly, the final annotations are passed through the post-processing phase to complete the pipeline. As annotating each SDOH type in separate prompts continued to yield higher recall than annotating all types in a single prompt, we tested both approaches with the new pipeline. With these enhanced configurations, generating annotations for each SDOH type in separate prompts fully outperformed generating all annotations with a single prompt (micro-F1 = 0.929 vs 0.890). Aggregating responses in this way enhances the model's performance by addressing variability in generated responses through self-consistency and by leveraging the advantages of a more specific prompt. 



\begin{table}[tb]
    \centering
    \caption{Comparison of few-shot learning on a subset of 30 from the MIMIC-III test set.}
    \label{tab:fewshot}
    \begin{tabular}{rccc}
\toprule
$n$-shot & Precision & Recall & Micro-F1\\
\midrule
0 & 0.764 & 0.810 & 0.786\\
10 & 0.827 & 0.821 & 0.824\\
30 & 0.866 & 0.879 & 0.872\\
50 & 0.888 & 0.875 & 0.881\\
100 & 0.744 & 0.768 & 0.756\\
\bottomrule
    \end{tabular}
\end{table}

\begin{table}[tb]
    \centering
    \caption{Comparison of self-consistency and post-processing on a subset of 30 from the MIMIC-III and UW test sets.}
    \label{tab:selfconsistency}
    \begin{tabular}{lccc}
\toprule
Prompt & Precision & Recall & Micro-F1\\
\midrule
Few-shot prompt & 0.802 & 0.885 & 0.842\\
+ self-consistency & 0.886 & 0.920 & 0.903\\
+ self-consistency \& post-processing & 0.948 & 0.911 & 0.929\\
\midrule
Generating all SDOH types at once & 0.884 & 0.896 & 0.890\\
\bottomrule
    \end{tabular}
\end{table}

\subsection{Error Analysis}

\begin{figure}[h]
    \centering
    \begin{subfigure}[b]{.4\linewidth}
    \centering
    \frame{\includegraphics[width=\linewidth]{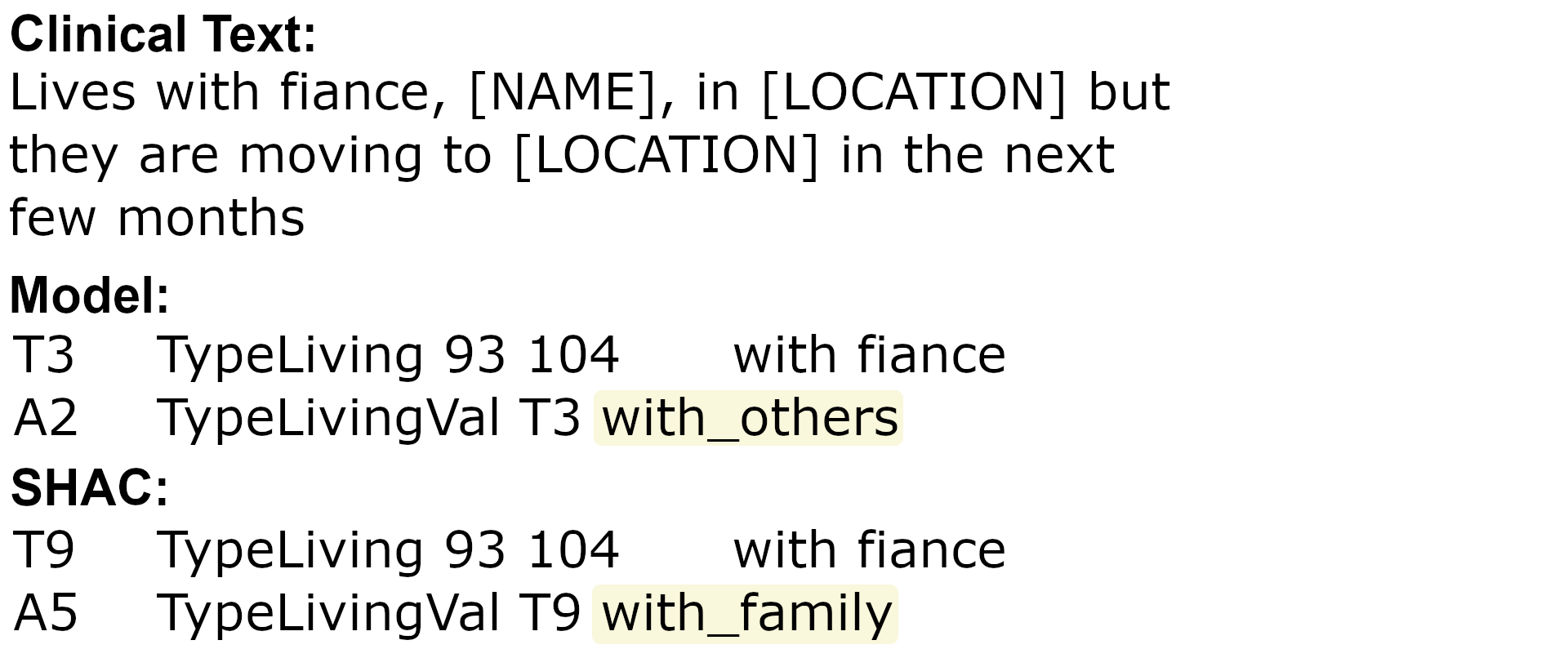}}
    \caption{\label{fig:sample1}}
    \end{subfigure}
    \begin{subfigure}[b]{.4\linewidth}
    \centering
    \frame{\includegraphics[width=\linewidth]{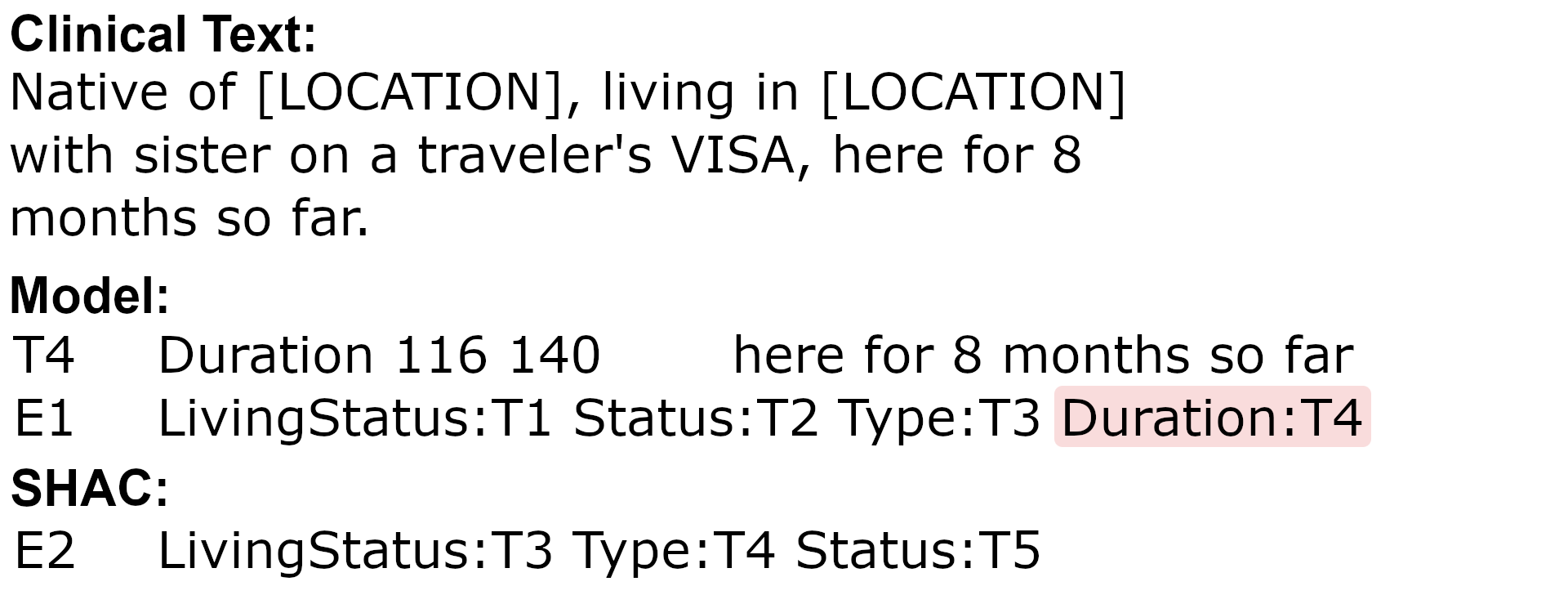}}
    \caption{\label{fig:sample2}}
    \end{subfigure}
    \begin{subfigure}[b]{.4\linewidth}
    \centering
    \frame{\includegraphics[width=\linewidth]{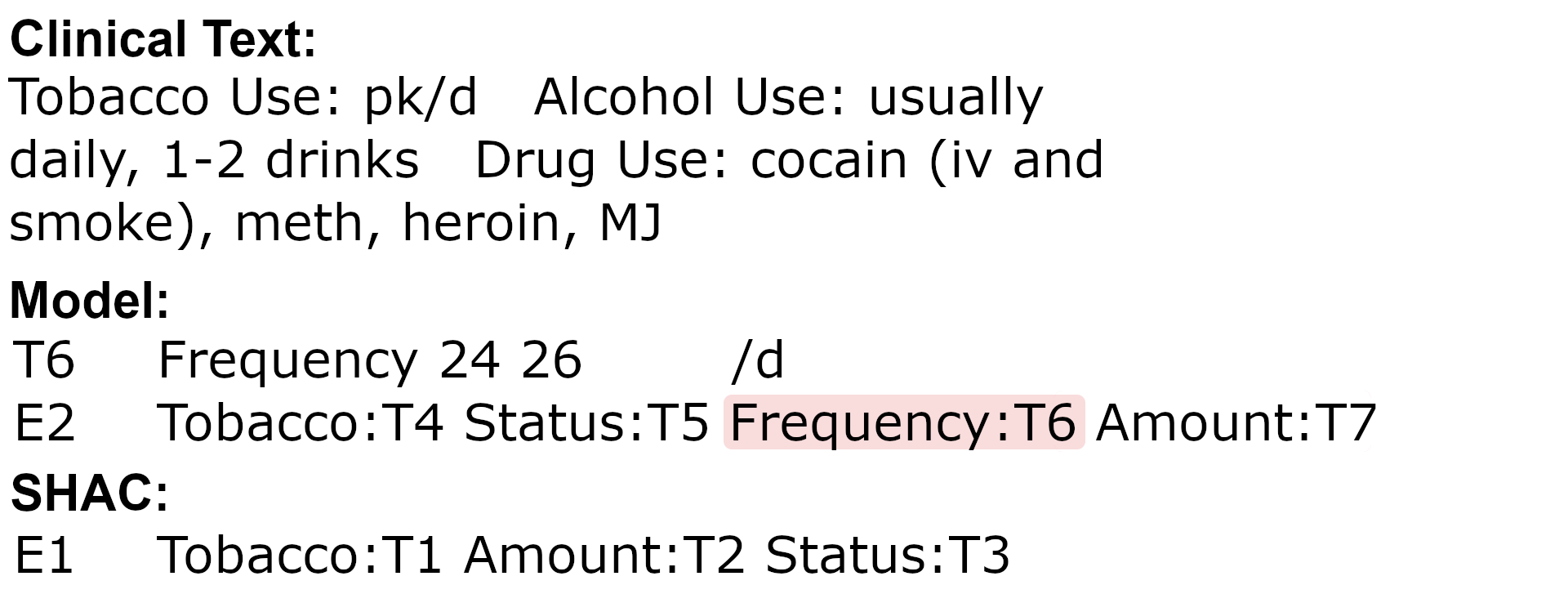}}
    \caption{\label{fig:sample3}}
    \end{subfigure}
    \begin{subfigure}[b]{.4\linewidth}
    \centering
    \frame{\includegraphics[width=\linewidth]{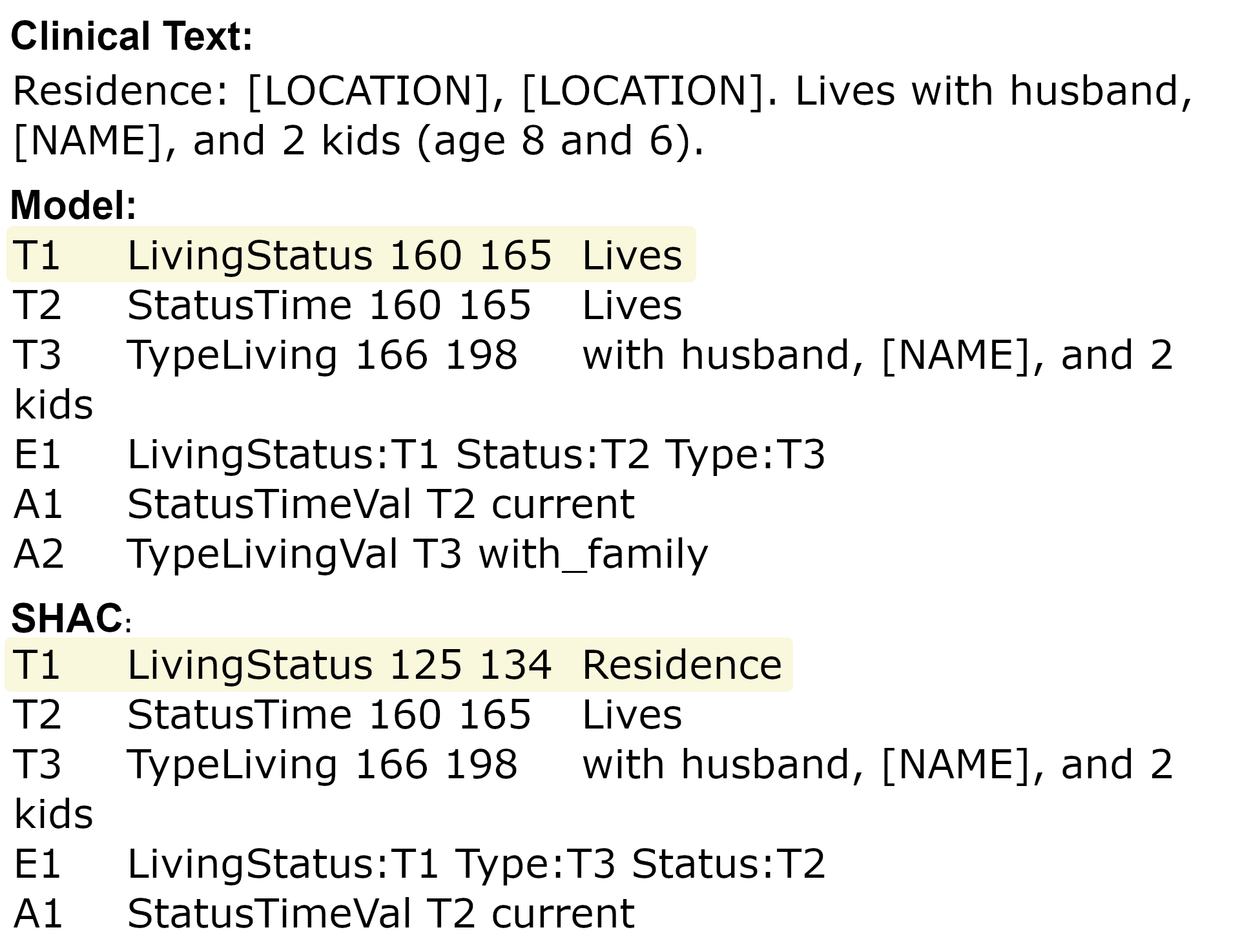}}
    \caption{\label{fig:sample4}}
    \end{subfigure}
    
    \caption{A sample of reasonably correct annotations that were marked as errors because they differed from the gold annotations.}
    \label{fig:keyerrors}
    \vspace{0.5em}
\end{figure}

To identify the areas in which our method encountered difficulties, we randomly selected a sample of 100 patient records, each with its corresponding prediction. We manually reviewed the model's results against the gold-standard annotations. We identified 78 errors that were penalized by the scoring algorithm (Table \ref{tab:error}). Notably, 23 of these errors were key-related issues rather than actual model errors, demonstrating that the model's true performance was likely higher than the recorded F1 score. Figure \ref{fig:keyerrors} shows a sample of annotations generated by the model that were marked as errors. Each example presents a clinical text excerpt, the model predictions, and the corresponding gold annotations from the SHAC corpus. In the first example, the model and the SHAC annotations differ in how they define living with a fianc\'{e} as ``with\_others'' or ``with\_family''.  The second and third examples demonstrate two cases in which the model adhered to the annotation guidelines and annotated all attributes, including those for which the gold annotations were missing. The attributes not in the gold annotations that the model identified were penalized by the scoring algorithm, yet they are verifiable through manual inspection. The fourth example illustrates a prediction difference: the model selected a different word as the trigger for LivingStatus than the gold annotations. We refer to these errors as key-related issues, and all the errors identified in this category largely followed these patterns, indicating a higher true performance for the model. Among the 55 observed model errors, the most frequent patterns involved the model's tendency to extract only one event per SDOH per record and its failure to synthesize events spanning longer sentence structures or to utilize contextual information. The model relied on explicit references. For example, it recognized ``homeless'' as a living status only when it was explicitly stated, ignoring phrases such as ``on the streets''. This demonstrates that LLMs could benefit from in-context learning for specific tasks, and that targeted tuning of LLM models could result in even higher measurable performance. The second row in the dataset under ``Errors by the model" corresponds to the instances where the model extracted non-specific phrases such as ``often," "rarely," or ``in the past," which were penalized by the scoring algorithm. These mistakes are also present in a small number of the gold annotations, therefore contributing to the model's predictions of similar phrases in this category despite the prompt's explicit instructions against it.

\begin{table}[tb]
    \centering
    \caption{Categories of errors in a sample of 100 predictions (50 MIMIC-III / 50 UW)}
    \label{tab:error}
    \small
    \begin{tabularx}{.7\linewidth}{Xr}
\toprule
Error Type & Freq\\
\midrule
Errors by the model & 55\\
~~Only included one event of many with the same SDOH type & 21\\
~~Non-specific type, frequency, history, duration, amount & 8\\
~~Missing an inferable but not explicit event & 7\\
~~Missing values in an extracted event & 6\\
~~Missing method / type for drug event & 5\\
~~Incorrect values due to not recognizing context & 4\\
~~Other & 4\\
Errors of the key / scoring (should not penalize the model) & 23\\
\bottomrule
    \end{tabularx}

\end{table}

\section{Discussion}

Our experiments demonstrate that LLMs with reasoning capabilities are strong contenders for SDOH extraction from clinical notes. LLMs also have several advantages over standard pre-trained LMs and other approaches. The primary advantage is their simplicity of implementation and ease of deployment. With a single prompt, independent researchers can consistently replicate results without requiring extensive training or fine-tuning of LMs. Second, LLMs demonstrate superior generalizability. Unlike traditional pre-trained models, LLMs can recognize SDOH information beyond their original scope without additional task-specific training. Finally, LLMs experience continuous improvement through iterative model releases. Emerging models, such as OpenAI’s GPT-5 and o3, have the potential to surpass previous generations, suggesting sustained advancements in SDOH extraction performance as the underlying technology evolves.

Recent studies have also demonstrated that LLMs struggle to understand longer prompts, often ignoring the body and focusing instead on the beginning and end. Strategies such as chain-of-thought prompting can be leveraged to further improve performance, mitigating this challenge by providing clear reasoning steps within concise prompts. Through further improvements and strategies, such as few-shot prompting and self-consistency, LLMs with reasoning capabilities are well-suited to extracting SDOH from clinical notes. 

Our study has several limitations. First, the data were drawn exclusively from the MIMIC-III and UW datasets, which were collected over a limited temporal and institutional range. This may limit the generalizability of our findings to other clinical scenarios. Second, the dataset only included five SDOH types and did not evaluate extraction performance on other common or less common SDOH. This limitation is shared by many studies in this domain, further constraining generalizability. Combatting this issue requires manual annotation of larger datasets for less common SDOH, which are difficult to recognize. This requires time and resources. 
Finally, our method relies on publicly available LLMs from commercial providers such as OpenAI. Prompting these models with clinical texts introduces potential privacy risks associated with the Protected Health Information (PHI). In practice, clinical texts processed using this method would be deidentified, minimizing any risk to patient privacy.

Our results are consistent with the state of the art on the same task achieved by pre-trained LMs. In addition, our error analysis indicates that the actual results may yield a higher F1-score than the measured one, and that effective prediction may be sufficient for downstream tasks in patient care and decision support.

\section{Conclusion}

In conclusion, LLMs with reasoning capabilities are strong contenders for extracting SDOH events from clinical corpora. Our experiments demonstrated that few-shot prompting substantially enhanced model performance, while the inclusion of a self-consistency module significantly reduced response variability, ensuring reliable and consistent annotations. These strategies, combined with optimized prompting and post-processing, produced results competitive with those of the top-performing LMs trained explicitly for this task, and demonstrated greater generalizability for less common SDOH than pre-trained LMs. Additionally, LLMs with reasoning capabilities require minimal setup and can be readily adopted by researchers. Finally, while our experiments focused on extracting SDOH events from the SHAC corpus, the proposed methodology applies to other NLP tasks and annotated corpora, offering broad prospects for clinical text processing.

\section*{Acknowledgment}

Research reported in this work was partially funded through a Patient-Centered Outcomes Research Institute (PCORI) Award (ME-2023C3-35934), National Library of Medicine grant (R01LM014306, R01LM014573), and National Science Foundation Graduate Research Fellowship under Grant No. 2139899.

\bibliographystyle{unsrtnat}
\bibliography{main}

\appendix


\begin{prompt}[Annotating SDOH]
\label{prompt:annotatingsdohfull}
\begin{lstlisting}[language=prompt]
Using the uploaded annotation guidelines for clinical notes, annotate the following clinical note from a doctor about a patient for the social determinants of health present in the notes. Output a list of events formatted in json format like this:
\end{lstlisting}

\begin{lstlisting}[language=json, escapechar=\%]
[{
  "sdoh": "[Employment|LivingStatus|Alcohol|Tobacco|Drug]",
  "trigger": (start_index, end_index, "text"),
  "status": (start_index, end_index, "text", "value") or null,
  "duration": (start_index, end_index, "text") or null,
  "history": (start_index, end_index, "text") or null,
  "type": (start_index, end_index, "text", "value") or null,
  "amount": (start_index, end_index, "text") or null,
  "frequency": (start_index, end_index, "text") or null,
  "method": (start_index, end_index, "text") or null
}]
\end{lstlisting}

\begin{lstlisting}[language=prompt]
Only include a field if it is present in the notes. Only include a value if it is present in the text. All sdoh events require the status field, and if the event is present the status is present in the text. Drug method refers to how the drug is used, and drug type refers to which drug is used (For example, IV/intravenous is a method, heroin is a type). If two or more annotations refer to the same physical event, then they must be combined to be one annotation. History, frequency, amount, and duration must be exact values, not "a lot," "often," "long time ago," etc. For employment trigger, prioritize words that are adjacent to "occupation" or "job" over job descriptions. 

Sometimes the input text is blank. Simply return an empty list of annotations.

Only return the output list. Add no comments, citations, or extra notes.

Here are some examples:
\end{lstlisting}

\begin{lstlisting}[language=json, escapechar=\%]
    %50 Annotated Examples Listed Here
\end{lstlisting}

\begin{lstlisting}[language=prompt]
Here are the notes to annotate:
\end{lstlisting}

\begin{lstlisting}[language=json, escapechar=\%]
    A note sample is shown in Figure %\ref{fig:brat}%
\end{lstlisting}

\begin{lstlisting}[language=prompt]
Be sure that the triggers for Tobacco events include words related to tobacco, smoking, chewing tobacco, or other uses.
Be sure that the triggers for Alcohol events include words related to alcohol, ethanol, or drinking
Be sure that the triggers for Drug events include words related to drugs or drug use
Be sure that the triggers for LivingStatus events include words related to living, residence, or location in hospitals/rehab facilities.
Be sure that the triggers for Employment events include words related to "work," "job," or "occupation" or words describing a title or field of work. Prioritize words related to "work," "job," or "occupation." If none are found, a job title is sufficient to be an employment trigger.

One event will be within one sentence unless another sentence doesn't have a sufficient trigger. If there is a trigger in both sentences, they are two separate sdoh events, and you should annotate them separately.
Only annotate a LivingStatus event if it can be said specifically if a person lives alone, with_others, with_family, or is homeless. If it is unclear, do not annotate it.
Tobacco can have a type: EX: cigarettes, pipe, chewing tobacco, etc.
Illicit drugs are a drug type
EmploymentStatus's value must be employed, unemployed, retired, student, or homemaker. 
Duration, Frequency, History, and Amount must be exact values (weekly, 30 years ago, per day, etc.), not references or generalizations (often, past, during WW2). 
Only TypeLiving requires a value in addition to extracted_text. Alcohol, Drug, Tobacco, and Employment type should only be extracted text, no value.
LivingStatus requires a type field. If it is not present, do not annotate the event.
Duration refers to how long an event persisted while History refers to how long ago an event ended (only applicable for past events).
One SDOH type will only have one event in a sentence. Two events of the same SDOH type cannot be in the same sentence.
The 'last use' of a substance refers to the most recent use and means it is currently in use.
\end{lstlisting}

\end{prompt}

\begin{prompt}[Annotating Each SDOH Type Separately]
\label{prompt:separatesdohfull}
\begin{lstlisting}[language=prompt]
Using the uploaded annotation guidelines for clinical notes, annotate the following clinical note from a doctor about a patient for the social determinants of health present in the notes. Output a list of events formatted in json format like this:
\end{lstlisting}

\begin{lstlisting}[language=json, escapechar=\%]
[{
  "sdoh": "[Employment|LivingStatus|Alcohol|Tobacco|Drug]",
  "trigger": (start_index, end_index, "text"),
  "status": (start_index, end_index, "text", "value") or null,
  "duration": (start_index, end_index, "text") or null,
  "history": (start_index, end_index, "text") or null,
  "type": (start_index, end_index, "text", "value") or null,
  "amount": (start_index, end_index, "text") or null,
  "frequency": (start_index, end_index, "text") or null,
  "method": (start_index, end_index, "text") or null
}]
\end{lstlisting}

\begin{lstlisting}[language=prompt]
Only include a field if it is present in the notes. Check for the trigger (required), status (required), type (required for LivingStatus), duration, frequency, history, method, and amount. 

Sometimes the input text is blank or no events are found. Simply return an empty list of annotations.

Only return the output list. Add no comments, citations, or extra notes.

Extract all of the {sdoh} events in the text. Do not extract any other sdoh type. Do not miss any of the {sdoh} events.

Here are some examples:
\end{lstlisting}

\begin{lstlisting}[language=json, escapechar=\%]
    %50 Annotated Examples Listed Here
\end{lstlisting}

\begin{lstlisting}[language=prompt]
Here are the notes to annotate:
\end{lstlisting}

\begin{lstlisting}[language=json, escapechar=\%]
    A note sample is shown in Figure %\ref{fig:brat}%
\end{lstlisting}

\begin{lstlisting}[language=prompt]
Be sure that the triggers for Tobacco events include words related to tobacco, smoking, chewing tobacco, or other uses.
Be sure that the triggers for Alcohol events include words related to alcohol, ethanol, or drinking
Be sure that the triggers for Drug events include words related to drugs or drug use
Be sure that the triggers for LivingStatus events include words related to living, residence, or location in hospitals/rehab facilities.
Be sure that the triggers for Employment events include words related to "work," "job," or "occupation" or words describing a title or field of work. Prioritize words related to "work," "job," or "occupation." If none are found, a job title is sufficient to be an employment trigger.

One event will be within one sentence unless another sentence doesn't have a sufficient trigger. If there is a trigger in both sentences, they are two separate sdoh events, and you should annotate them separately.
Only annotate a LivingStatus event if it can be said specifically if a person lives alone, with_others, with_family, or is homeless. If it is unclear, do not annotate it.
Tobacco can have a type: EX: cigarettes, pipe, chewing tobacco, etc.
Illicit drugs are a drug type
EmploymentStatus's value must be employed, unemployed, retired, student, or homemaker. 
Duration, Frequency, History, and Amount must be exact values (weekly, 30 years ago, per day, etc.), not references or generalizations (often, past, during WW2). 
Only TypeLiving requires a value in addition to extracted_text. Alcohol, Drug, Tobacco, and Employment type should only be extracted text, no value.
LivingStatus requires a type field. If it is not present, do not annotate the event.
Duration refers to how long an event persisted while History refers to how long ago an event ended (only applicable for past events).
One SDOH type will only have one event in a sentence. Two events of the same SDOH type cannot be in the same sentence.
The 'last use' of a substance refers to the most recent use and means it is currently in use.
\end{lstlisting}

\end{prompt}

\renewcommand{\tablename}{Supplementary Table}
\renewcommand{\thetable}{\Roman{table}}
\setcounter{table}{0}

\begin{table}[!htbp]
\centering
\caption{Performance of Llama-3.1-8b model on n2c2 shared task.}

\label{tab:results_appendix}
\begin{tabular}{lrrr}
\toprule
 & Precision & Recall & Micro-F1\\
\midrule
Our method (llama-3.1-8b) & 0.577 & 0.605 & 0.591\\
~~Alcohol & 0.616 & 0.635 & 0.626 \\
~~Drug & 0.568 & 0.452 & 0.503 \\
~~Employment & 0.382 & 0.537 & 0.446 \\
~~Living Status & 0.690 & 0.757 & 0.722 \\
~~Tobacco & 0.566 & 0.598 & 0.582 \\
\bottomrule
\end{tabular}
\end{table}

\end{document}